\documentclass[conference]{IEEEtran}
\IEEEoverridecommandlockouts
\usepackage{cite}
\usepackage{amsmath,amssymb,amsfonts}
\usepackage{algorithmic}
\usepackage{graphicx}
\usepackage{textcomp}
\usepackage{xcolor}

\usepackage{textcomp}
\usepackage{stfloats}
\usepackage{url}
\usepackage{verbatim}
\usepackage{graphicx}
\usepackage{tabularray}
\usepackage{booktabs,makecell, multirow, tabularx} 
\usepackage{caption}
\usepackage{subcaption}
\usepackage{adjustbox,lipsum}

\def\BibTeX{{\rm B\kern-.05em{\sc i\kern-.025em b}\kern-.08em
    T\kern-.1667em\lower.7ex\hbox{E}\kern-.125emX}}
\begin{document}

\title{TransGlow: Attention-augmented Transduction model based on Graph Neural Networks for Water Flow Forecasting

\author{Naghmeh Shafiee Roudbari, Charalambos Poullis, Zachary Patterson, Ursula Eicker \\
    \textit{Gina Cody School of Engineering and Computer Science} \\
    \textit{Concordia University}\\
    \textit{Montreal, QC, Canada}
    }
% *\\
}

% \and
% \and
% \and
% \and
% \and

\maketitle

\begin{abstract}
The hydrometric prediction of water quantity is useful for a variety of applications, including water management, flood forecasting, and flood control. However, the task is difficult due to the dynamic nature and limited data of water systems. Highly interconnected water systems can significantly affect hydrometric forecasting. Consequently, it is crucial to develop models that represent the relationships between other system components. In recent years, numerous hydrological applications have been studied, including streamflow prediction, flood forecasting, and water quality prediction. Existing methods are unable to model the influence of adjacent regions between pairs of variables.  In this paper, we propose a spatiotemporal forecasting model that augments the hidden state in Graph Convolution Recurrent Neural Network (GCRN) encoder-decoder using an efficient version of the attention mechanism. The attention layer allows the decoder to access different parts of the input sequence selectively. Since water systems are interconnected and the connectivity information between the stations is implicit, the proposed model leverages a graph learning module to extract a sparse graph adjacency matrix adaptively based on the data. Spatiotemporal forecasting relies on historical data. In some regions, however, historical data may be limited or incomplete, making it difficult to accurately predict future water conditions. Further, we present a new benchmark dataset of water flow from a network of Canadian stations on rivers, streams, and lakes. Experimental results demonstrate that our proposed model TransGlow significantly outperforms baseline methods by a wide margin.
\end{abstract}


\section{Introduction}
Accurate water flow prediction plays a crucial role in flood forecasting and mitigation. By understanding and predicting the dynamics of water flow, authorities can issue timely warnings and implement proactive measures to minimize the impact of floods, protecting human lives and reducing property damage. This proactive approach allows for better emergency response planning and the implementation of effective flood control strategies. Furthermore, water flow prediction is essential for optimal water resource management, fair distribution of water, ensuring sustainable use, and minimizing waste. 

Water systems are interconnected with interdependencies, which can significantly impact hydrometric prediction. Water levels, flow, and quality changes in one part of the system can have cascading effects on the other parts. For example, changes in precipitation in one part of a river basin can affect water levels and flows downstream. These dependencies are challenging to understand, as different components can interact in complex ways that rely on various factors. Hence, it is essential to develop models that can capture the relationships between other system components. Spatiotemporal forecasting in water flow prediction involves capturing the complex relationships and patterns of water flow in a given geographical area over time. It takes into account the interconnections of hydrological processes across different locations and time intervals. The concept of spatiotemporal forecasting recognizes that water flow is not only influenced by local conditions but also by the spatial context and interactions within the hydrological system. It considers how changes in one area can propagate and affect water flow patterns in neighboring or downstream locations. Additionally, it takes into account the temporal dynamics, such as seasonality, trends, and short-term variations, that influence water flow.

Previous studies in the field of hydrometric prediction can be categorized into distinct research areas. These categories include streamflow forecasting \cite{kicsi2008stream}, drought prediction \cite{hao2018seasonal,hao2014global,hao2017overview}, flood forecasting \cite{nayak2005short,jain2018brief,karyotis2019deep}, and water quality prediction \cite{ahmed2019machine,haghiabi2018water,mohr2021assessment}. Previous studies in hydrometric prediction have significantly contributed to the field; however, they are not without limitations. These studies have often relied on limited and fragmented datasets, which can result in uncertainties and reduced accuracy. Additionally, oversimplified assumptions about hydrological processes and the disregard of spatial and temporal variability can reduce the accuracy of the predictions. Furthermore, limited focus on realtime applications poses challenges for the field. 

To address the mentioned challenges, we propose TransGlow, a spatiotemporal forecasting solution based on a transductive model with an augmented decoder hidden state using an efficient attention mechanism. The attention ability to focus on relevant parts of the input allows the model to reduce the risk of losing context information from the beginning of the sequence. In time series modeling, it is necessary to preserve the ordering information. However, the permutation-invariant self-attention mechanism results in temporal information loss\cite{zeng2023transformers}. Recurrent neural network (RNN) family \cite{medsker2001recurrent,graves2012long,dey2017gate} have been well known as state of the art approaches in sequence modeling and transduction problems such as language modeling and machine translation \cite{mikolov2012context,datta2020neural}. We use attention in parallel with a vanilla RNN-based encoder to maintain ordering information and capture more relevant contextual information simultaneously. With this approach, the information can be spread throughout the RNN encoder and the attention layer, then selectively retrieved by the decoder allowing the model to process sequential data effectively. The main weakness of attention mechanism is the high computational complexity and memory usage while computing the dot product. To avoid this, we use ProbSparse self-attention, an efficient attention mechanism proposed by \cite{zhou2021informer}.

The purpose of Graph Neural Networks (GNNs) \cite{hamilton2020graph} is to analyze data represented as graphs. They can operate on nodes and edges, capturing the graph's structural information and relationships. We can leverage the power of GNNs by incorporating Graph Convolution operations  \cite{kipf2016semi} into the encoder-decoder design; therefore, we employ Graph Convolution Recurrent Neural Networks \cite{li2017diffusion} blocks to capture spatial dependencies and extract high-level representations for each water station. The self-learning graph structure has numerous benefits. It captures the changing relationships among variables over time, allowing the model to adapt to shifting patterns and dependencies. This dynamic graph construction is especially useful when the relationships between variables are ambiguous or change over time intervals. The model can learn to establish relationships between relevant variables based on their temporal dependencies and the current forecasting task. This adaptability enables TransGlow to capture both local and global dependencies between variables, resulting in more accurate forecasting. Our principal contributions are as follows:

\begin{enumerate}
    \item An augmented transduction model with an efficient attention mechanism for spatiotemporal forecasting.
    
    \item To the best of our knowledge, this is the first study of water flow forecasting from a graph-based perspective to learn the actual correlation between drainage basins.

    \item We present our experiments on 186 drainage basins across Canada. The raw data is provided by the Environment and natural resources of Canada \cite{Canada_2018}. Experimental results show that our method outperforms the state-of-the-art methods on all prediction horizons and performance metrics. Detailed descriptions of the datasets are provided in Section \ref{sec:experiments}.
\end{enumerate}

The rest of the paper is organized as follows. Section \ref{sec:literature} reviews the current state of the literature in spatiotemporal forecasting. Section \ref{sec:model} formally states the problem and the proposed methodology. Finally, Section \ref{sec:experiments} includes dataset description, experimental settings, and experiment results on a real-world dataset to show the effectiveness of our proposed approach.

\section{Related Work}
\label{sec:literature}

A plethora of data-driven methods have been proposed for Time series Forecasting (TSF) in the past. The prevailing focus in the literature is on statistical approaches, with notable methods like Vector Auto Regressive models \cite{zivot2006vector}, Autoregressive Integrated Moving Average (ARIMA)\cite{lai2020use}, and its variants \cite{narasimha2018modeling}. While statistical models have gained popularity due to their simplicity and interpretability, they heavily rely on assumptions related to stationary processes, which may not always hold true in real-world scenarios, especially for multivariate time series data. On the other hand, machine learning methods demonstrated a solid ability to learn nonlinearity in TSF. \cite{kicsi2008stream} developed a wavelet model for stream flow prediction. For water quality prediction, \cite{ahmed2019machine} proposed a Support Vector Machine (SVM) approach, and \cite{haghiabi2018water} proposed a Multi-Layer Perepcetron (MLP), the results showed that the model has good predictive performance compared to other baselines. All these studies demonstrate strong capabilities of machine learning for complex nonlinear feature extraction and improve prediction accuracy. However, limitations in capturing spatial and temporal dependencies for more complex data prompted the exploration of deep learning methods. 

Recurrent Neural Networks (RNNs) \cite{medsker2001recurrent} with internal memory became an excellent choice for time-series forecasting, particularly Long Short-Term Memory (LSTM) \cite{graves2012long} and Gated Recurrent Unit (GRU) \cite{dey2017gate} models with the ability to address vanishing gradient problems and effectively learning longterm temporal dependency. By introducing attention mechanisms \cite{vaswani2017attention}, different Transformer based methods have also been applied to TSF applications such as traffic \cite{xu2020spatial}, air quality \cite{liang2023airformer}, and energy \cite{zhou2021informer} forecasting. However, a recent study \cite{zeng2023transformers} claims that Transformers are not effective for time series forecasting by comparing the Transformer-based models against simple one-layer linear models. To validate this claim, we also conduct experiments to explore the impacts of transformer-based models in TSF; detailed descriptions are provided in Section \ref{sec:experiments}.

% Numerous RNN-based models, such as bidirectional LSTM \cite{siami2019performance}, Mixture Deep LSTM \cite{yu2017deep}, and GRU models \cite{lin2022time}, were applied to TSF problems with impressive results

To capture spatial dependencies, Convolutional Neural Networks (CNNs) were introduced, treating TSF data as a time-space matrix \cite{ma2017learning}. However, CNN models are limited to grid-like structures and Euclidean space representation. Then, Graph Neural Networks (GNNs) came into the spotlight, offering a powerful way to express complex relationships in unstructured data using graph-based data structure. Graph-based methods have been widely used in different spatiotemporal applications such as solar energy, traffic, and electricity \cite{wu2020connecting}. Still, their application in predicting hydrological-related parameters and water resources is relatively limited. Recent studies combined GNNs with RNNs to explore spatial and temporal changes \cite{bai2020adaptive}, achieving promising results. Encoder-Decoder architectures have shown great potential for processing sequence data \cite{makin2020machine}. Researchers applied this architecture to TSF,  using GCRN \cite{li2017diffusion}, attention \cite{zheng2020gman}, and transformer-based architectures \cite{xu2020spatial}. Here we propose a novel transduction architecture using the attention mechanism to augment the hidden state and enable better information flow and context preservation. To utilize the strong potential of RNN models in capturing temporal information and GNN in uncovering spatial relationships, we use GCRN as the building blocks of our core Encoder-Decoder model. For the graph convolution operation in GCRN, we employ a self-learning graph module to adaptively understand implicit spatial dependencies.

\begin{figure*}[!t]
    \centering
    \includegraphics[width=\textwidth]{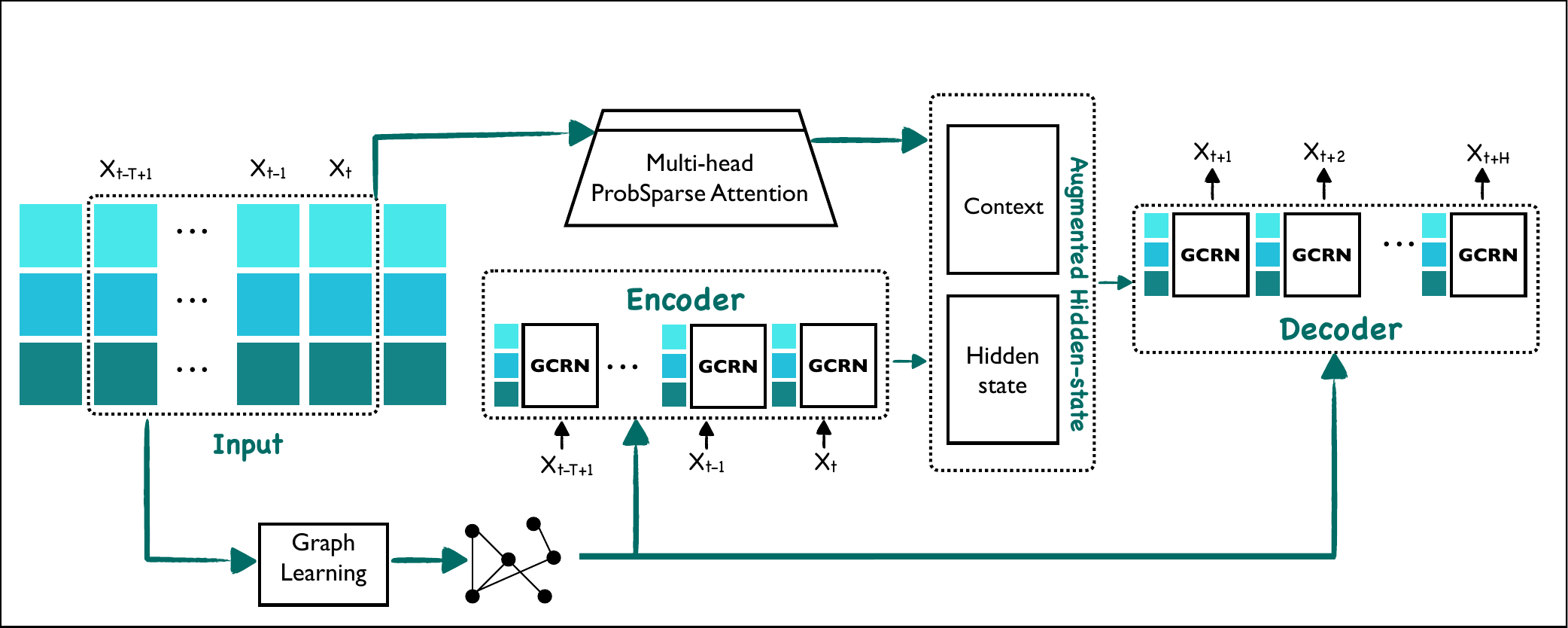}
    \caption{TransGlow main Architecture: graph learning module learns the graph to be used in GCRN blocks. The final hidden state of the Encoder is augmented by the attention context vector to pass over to the Decoder}
\label{fig:architecture}
\end{figure*}

\section{Methodology}
\label{sec:model}
\subsection{Problem Statement}

The spatiotemporal input dataset containing water flow measurements at different monitoring stations over time is given as a two-dimensional tensor $X \in \mathbb{R}^{n\times d}$, where $n$ refers to the total number of water stations generating their own time series, and $d$ refers to the total number of time intervals covered by the dataset. $X^t \in \mathbb{R}^{n\times T}$ denotes a sequence of historical data from time $t-T+1$ to time $t$ over all the $n$ resources.

\begin{center}
% $f: {X}^t, {A} \rightarrow {Y}$
% \\
\vspace{.15cm}
$X^t =
\begin{bmatrix}
x_0^{t-T+1} & \dots & x_0^{t-1} & x_0^{t} \\
x_1^{t-T+1} & \dots & x_1^{t-1} & x_1^{t} \\
\dots & \dots & \dots & \dots \\
x_{n-1}^{t-T+1} & \dots & x_{n-1}^{t-1} & x_{n-1}^{t} \\
\end{bmatrix}_{n*T}$
\end{center}
\vspace{.15cm}

Here, the goal is to predict future tensor values up to $H$ steps ahead ($X^{t+H}$) given historical data from the last T steps ($X^{t}$). The methodology involves improving accuracy by taking spatial dependency between all water stations ($n$ nodes) into account through graph convolution operation. Since the graph structure is unknown, the graph adjacency matrix $A_{n\times n}$ needs to also be extracted. So, the objective is to find a function $f$  that learns the graph as $A$ from input sequence $X^{t}$:
\begin{center}
$f: {X}^t \rightarrow {A}$
% \\
\end{center}
\vspace{.15cm}
then find a function $g$ that maps the input $X^{t}$ and $A$ to the future values:
\begin{center}
$g: {X}^t, {A} \rightarrow {X}^{t+H}$
% \\
\end{center}
\vspace{.15cm}

\subsection{Graph Learning Module}

The underlying graph structure for graph convolution operation is either defined based on distance and similarity functions or needs to be constructed. Having a predefined graph requires specific prior knowledge of the problem. The spatial relationship between the objects is often implicit, so even prior information might be biased and misleading for the prediction task. Several existing studies \cite{wu2020connecting},\cite{jiang2023spatio},\cite{bai2020adaptive} serve this purpose mostly as a function of the node embedding's product. Here we adopt the well-established adaptive graph generation defined in \cite{bai2020adaptive}:

\begin{equation}
\label{eq:graph_conv}
\hat{A}=  softmax(relu( E1 . E2^T))
\end{equation}
where E1 and E2 represent randomly initialized node embeddings, which are subject to learning during the training process.

\subsection{Graph Convolution Recurrent Block}

The blocks of the encoder-decoder architecture consist of two primary modules: a graph convolution operation to capture spatial dependencies and an RNN-based unit to exploit temporal variability. GCRN has been firmly established as state of the art approach in spatiotemporal forecasting problems. Building upon the widely used approach of combining graph convolution within an RNN layer in the literature \cite{li2017diffusion,jiang2023spatio,bai2020adaptive}, we adopt the original GCRN formulation initially proposed by \cite{li2017diffusion}, by eliminating the dual random walk. This modification is aimed at avoiding computational complexities. The graph convolution operation is a simplified version of normalized Laplacian formulated as:

\begin{equation}
\label{eq:graph_conv}
{H}^{(l)}= \sigma (\hat{A}H^{(l-1)}W^{(l)})+ b^l
\end{equation}
where $\hat{A}$ denotes the learned graph, $W^l$ and $b^l$ are the trainable weight and bias matrices for layer $l$, and $\sigma$ is the activation function. $H^{l-1}$ and $H^l$ are the input signal and output of graph convolution on layer $l$, respectively. Finally, the MLP layers of the Gated Recurrent Unit are replaced by the introduced graph convolution operation above to form the block:

\begin{equation}
\label{eq:rt}
r^{t}= \sigma (\mathcal{G}_{conv}(A,[X^t,h^{t-1}])+ b_r
\end{equation}
\begin{equation}
\label{eq:ut}
u^{t}= \sigma (\mathcal{G}_{conv}(A,[X^t,h^{t-1}])+ b_u
\end{equation}
\begin{equation}
\label{eq:ct}
c^{t}= tanh (\mathcal{G}_{conv}(A,[X^t,r^t *h^{t-1}])+ b_c
\end{equation}
\begin{equation}
\label{eq:ht}
h^{t}= (u^t * h^{t-1})(1.0-u^t)*c
\end{equation}
where $\mathcal{G}_{conv}$ is the graph convolution operation from Equation \ref{eq:graph_conv}, $r^t$ and $u^t$ are the reset gate and the update gate, $c^t$ is the cell state, $X^t$ is the input signal, $\sigma(.)$ and $tanh(.)$ are activation functions, and $h$ is the hidden state.

\subsection{Encoder-Decoder Model}
The flexibility of the encoder-decoder design and its effectiveness in handling sequence-to-sequence tasks have made it a popular choice across various domains, such as machine translation \cite{datta2020neural} and time series forecasting \cite{zhou2021informer}, enabling Neural Networks architectures to process and generate human-like data sequences. The vanilla encoder-decoder model works in a sequence-to-sequence manner. The encoder processes the input sequence to create a fixed-size representation that carries the relevant information. This context vector is then passed to the decoder, which uses it to generate the output sequence one step at a time. The model building blocks can be implemented using RNNs, LSTMs, or transformers.

The main problem of the vanilla encoder-decoder architecture is that it may suffer from the issue of information compression and loss. Since the encoder produces a fixed-size representation (context vector) to summarize the entire input sequence, it needs to capture all the relevant information within this fixed-size vector. However, this process can lead to lossly information compression, where important details from the input sequence may get lost or diluted in the context vector. Additionally, the encoder-decoder architecture may face difficulty in handling long sequences. When processing long input sequences, the encoder's fixed-size context vector may not be sufficient to retain all the essential information, resulting in inadequate generation of the output sequence by the decoder.
To address these problems, we propose a modification and improvement to the vanilla encoder-decoder architecture by using attention mechanisms to mitigate information loss, handle longer sequences more effectively, and improve the model's overall performance in spatiotemporal forecasting. The added layer between the input data and the decoder is a way of attending to the input sequence by an attention distribution mechanism that calculates a weighted sum of the inputs at all time steps. The final augmented hidden state $H$ passed over to the decoder is:

\begin{equation}
\label{eq:encoder2}
H= Concat[h^t,C]
\end{equation}
where $h^t$ is the final hidden state of the encoder, and $C$ is the context vector from the attention layer. The attention vector is then incorporated into the decoder's decision-making process, allowing it to focus on relevant information from the source data during decoding.  Figure \ref{fig:architecture} visually illustrates an encoder-decoder model with an augmented attention layer.

\subsection{Efficient Attention Layer}

The bottleneck of the original attention mechanism lies in its quadratic computational complexity with respect to the sequence length. \cite{zhou2021informer} have been able to alleviate the bottleneck of canonical attention and make Transformer-based models more scalable for sequence-to-sequence tasks by proposing ProbSparse Self-attention. Instead of having a full query matrix with all the queries for every token in the sequence, ProbSparse attention selects only a subset of k queries based on a certain measure or probability distribution:

\begin{equation}
\label{eq:prob_q}
\hat{Q}= M(Q,K)
\end{equation}

\begin{equation}
\label{eq:prob_att}
Attention(Q,K,V)= Softmax(\frac{\hat{Q}K^T}{\sqrt{d}})V
\end{equation}
where $Q$,$K$,$V$ denote query, key, and value, respectively, and $d$ is the input dimension. The measure or probability distribution $M$ determines the importance or relevance of each token in the sequence with respect to the current token. Tokens with higher importance or relevance are more likely to be included in the sparse query matrix, while tokens with lower importance have a lower probability of being included. Therefore, we utilize the ProbSparse attention mechanism for the attention augmented layer in the encoder-decoder design.

\section{Experiments}
\label{sec:experiments}
\subsection{Dataset Description}

\begin{figure}[!t]
    \centering
    \begin{subfigure}{0.44\textwidth}
    \includegraphics[width=\textwidth]{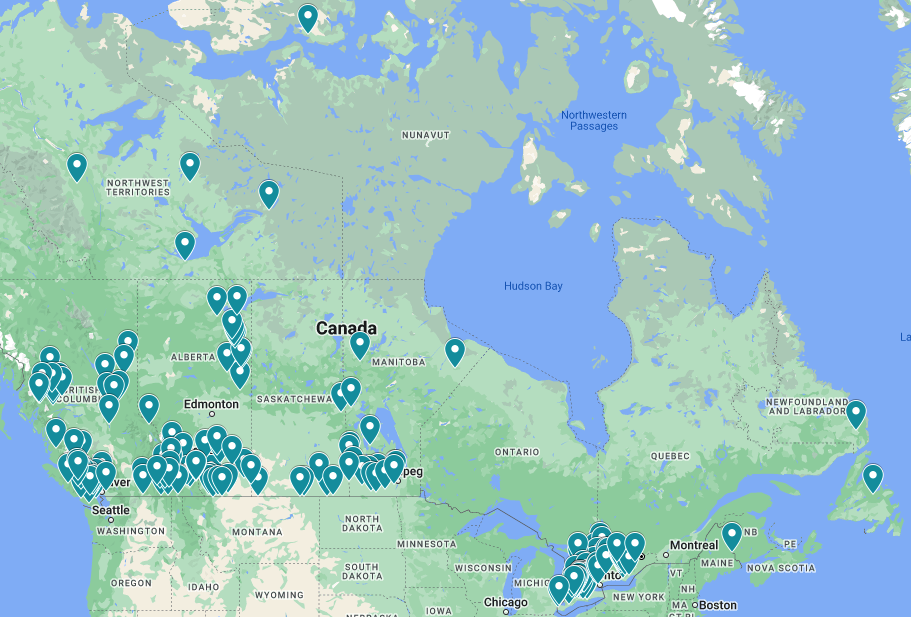}
            \caption{}
        \label{subfig:flow_coverage}
    \end{subfigure}
    % \hfill
    \begin{subfigure}{0.44\textwidth}    \includegraphics[width=\textwidth]{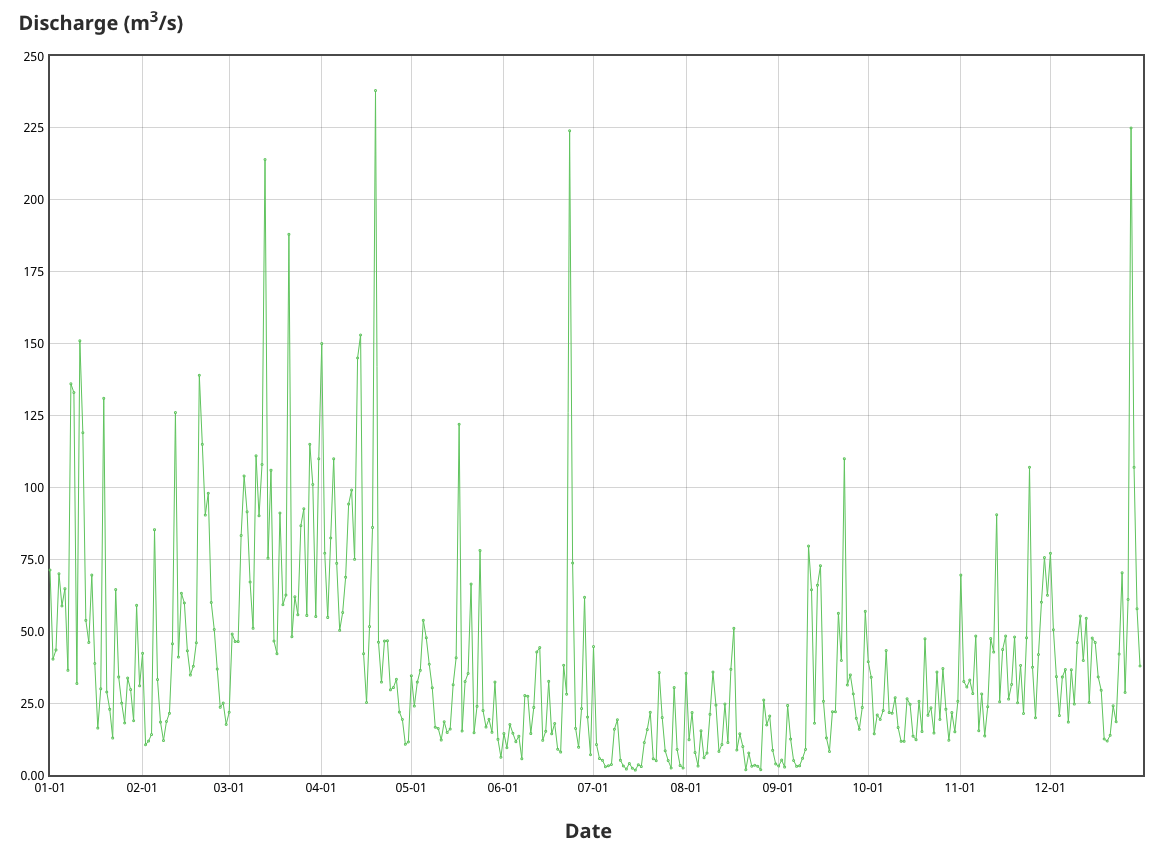}
            \caption{}
        \label{subfig:variablity}
    \end{subfigure}
    \caption{CWFDD-186 dataset (a) 186 station's distribution on the map. (b) Data variability pattern for a specific sensor during a year}
    \label{fig:maps2}
\end{figure}

In this work, we present a water flow daily discharge dataset called CWFDD-186, which provides daily discharge data from 186 stations on rivers, streams, and lakes across Canada. The raw data is provided by Environment and Climate Change Canada \cite{flowData}. The presented dataset, publicly available on the project repository, covers 40 years from 1981 to 2021. Figure \ref{subfig:flow_coverage} shows the station's coverage on the map, and \ref{subfig:variablity} is the daily discharge of one of the stations over a year period. 

The missing values are replaced using the Historical Average method, with the weighted average of previous/next years. Occasionally spikes or dips can occur in the data due to sensor malfunctions, data transmission errors, or calibration issues with the measurement instruments. Environmental factors such as debris, ice, or sudden changes in water flow conditions can also lead to temporary fluctuations in the data. We have applied a Gaussian smoothing filter on data to address this without losing the general pattern and trends of the data. Smoothed data helps the model catch actual patterns by removing the noise.

\subsection{Experimental Settings}
We use 70\% of the data for training, 10\% for
validation, and 20\% for testing with batch size 64. The historical sequence length and the prediction horizon are both set to 12. The maximum number of epochs is set to 200, while the training may stop earlier if validation converges for 20 consecutive epochs. We use the Adam optimizer for training with Mean Absolute Error (MAE) as the loss function and curriculum learning for better generalization. The base learning rate is 0.01,  the decay ratio is 0.1, and the number of heads for attention is 8. The network is implemented using PyTorch v1.7.1. All experiments are conducted on NVIDIA GeForce RTX 2080 Ti with 11GB memory. 

\subsection{Baselines}

We compare our proposed TransGlow method with various baseline models, including graph neural networks, transformers, attention-based, and statistical methods. The baseline models include the following:
\begin{enumerate}
  \item The Historical Average model (HA) is forecasting method that predicts the future value of a time series based on the weighted average of historical data points. 
  \item Vector Auto-Regression (VAR) is a statistical model used for multivariate time series forecasting. It is an extension of the univariate Auto-Regression (AR) model to handle multiple related time series variables simultaneously. VAR is widely used in econometrics, finance, and various other fields for analyzing and predicting the interactions between multiple variables over time.
  \item The Adaptive Graph Convolutional Recurrent Network (AGCRN) \cite{zhou2017graph} is designed to capture both temporal and spatial dependencies within graph-structured data effectively. AGCRN takes into account temporal information as intra-dependencies and spatial information as inter-dependencies. To achieve this, it integrates two crucial adaptive modules based on Recurrent Neural Networks (RNNs) and Graph Neural Networks (GNNs).
\item The Informer model \cite{zhou2021informer} is introduced as an enhanced version of the Transformer architecture specifically designed for efficient and accurate forecasting of long time-series sequences. The proposed modifications, including temporal attention and ProbSparse self-attention mechanisms, enable Informer to achieve state-of-the-art performance while efficiently handling long-range dependencies in time-series data.
\item Spatial-Temporal Transformer Networks for Traffic Flow Forecasting (STTransformer) \cite{xu2020spatial} introduces a novel application of the Transformer architecture by proposing spatial and temporal transformers to capture spatial and temporal dependencies in traffic flow data. The model achieves state-of-the-art performance in traffic flow forecasting tasks by treating the traffic network as a graph and utilizing multi-head attention mechanisms.
\item MTGNN \cite{wu2020connecting} uncovers the relationships between variables using a graph learning module. The model also incorporates a unique mix-hop propagation layer and a dilated inception layer. All three modules are jointly learned within an end-to-end framework, ensuring optimal integration of the learning process. 
\item Discrete Graph Structure (GTS) \cite{shang2021discrete} is a scalable spatiotemporal forecasting method where the number of parameters does not grow quadratically with the number of time-series. It was originally proposed for traffic forecasting and uses the GCRN block structure and a graph learning approach they present in the paper. 
\end{enumerate}

\subsection{Performance Comparison}
Table \ref{performance_table} shows a comparison of the performance for the above-mentioned methods. We report performance metrics for three different horizons: 3, 6, and 12 days.  Widely used measures in TSF literature are employed to report the performance, including (1) Mean Absolute Error (MAE), (2) Root Mean Squared
Error (RMSE).

\begin{equation}
	\label{eq:rmse_eval}
	RMSE = \frac{1}{n}\sum_{i=1}^{h}|\frac{Y^{t+i} - X^{t+i}}{Y^{t+i}}|
\end{equation}

\begin{table}
\caption{Experimental results of various baseline methods conducted on water flow data for different values of time steps ahead}
\label{performance_table}
\centering
\begin{tabular}{cc|ccc}

% { |p{3cm}||p{3cm}|p{3cm}|p{3cm}|  }
% {cc|ccc}
\hline
                       & Dataset & \multicolumn{3}{c}{CWFDD-186}                                                \\ \hline
                       &         & \multicolumn{3}{c}{Horizon}                                                \\ \hline
\multicolumn{1}{l}{Method}                 & \multicolumn{1}{c}{Metric}  &  \multicolumn{1}{c}{3 Days} &
\multicolumn{1}{c}{6 Days} & \multicolumn{1}{c}{12 Days}                                                      \\ \hline

\multirow{3}{*}{HA} & \multicolumn{1}{c|}{MAE}  & \multicolumn{1}{c}{44.82} & \multicolumn{1}{c}{44.82} & \multicolumn{1}{c}{44.82}  \\ &\multicolumn{1}{c|}{RMSE}  & \multicolumn{1}{c}{61.39} & \multicolumn{1}{c}{61.39} & \multicolumn{1}{c}{61.39}  \\ \hline

\multirow{3}{*}{VAR} & \multicolumn{1}{c|}{MAE} & \multicolumn{1}{c}{23.29} & \multicolumn{1}{c}{30.19} & \multicolumn{1}{c}{36.39} \\ & \multicolumn{1}{c|}{RMSE} & \multicolumn{1}{c}{94.17} & \multicolumn{1}{c}{131.76} & \multicolumn{1}{c}{173.55} \\ \hline

\multirow{3}{*}{AGCRN}  
& \multicolumn{1}{c|}{MAE}  & \multicolumn{1}{c}{11.27}    
& \multicolumn{1}{c}{13.64}   & \multicolumn{1}{c}{16.40}    \\ 
& \multicolumn{1}{c|}{RMSE}  & \multicolumn{1}{c}{44.57}    
& \multicolumn{1}{c}{46.30}   & \multicolumn{1}{c}{49.64}    \\ \hline

\multirow{3}{*}{Informer}  
& \multicolumn{1}{c|}{MAE}  & \multicolumn{1}{c}{17.68}    
& \multicolumn{1}{c}{17.93}   & \multicolumn{1}{c}{18.76}    \\ 
& \multicolumn{1}{c|}{RMSE}  & \multicolumn{1}{c}{44.71}    
& \multicolumn{1}{c}{45.51}   & \multicolumn{1}{c}{48.19}    \\ \hline

\multirow{3}{*}{STTransformer}  
& \multicolumn{1}{c|}{MAE}  & \multicolumn{1}{c}{9.44}    
& \multicolumn{1}{c}{11.03}   & \multicolumn{1}{c}{14.33}    \\ 
& \multicolumn{1}{c|}{RMSE}  & \multicolumn{1}{c}{27.20}    
& \multicolumn{1}{c}{30.85}   & \multicolumn{1}{c}{39.27}    \\ \hline

\multirow{3}{*}{MTGNN}  
& \multicolumn{1}{c|}{MAE}  & \multicolumn{1}{c}{7.98}    
& \multicolumn{1}{c}{9.47}   & \multicolumn{1}{c}{58.72}    \\ 
& \multicolumn{1}{c|}{RMSE}  & \multicolumn{1}{c}{24.04}    
& \multicolumn{1}{c}{27.59}   & \multicolumn{1}{c}{107.34}    \\ \hline

\multirow{3}{*}{GTS}  
& \multicolumn{1}{c|}{MAE}  & \multicolumn{1}{c}{7.33}    
& \multicolumn{1}{c}{9.93}   & \multicolumn{1}{c}{12.53}    \\ 
& \multicolumn{1}{c|}{RMSE}  & \multicolumn{1}{c}{22.56}    
& \multicolumn{1}{c}{28.97}   & \multicolumn{1}{c}{34.52}    \\ \hline

\multirow{3}{*}{TransGlow}  
& \multicolumn{1}{c|}{MAE}  & \multicolumn{1}{c}{6.83}    
& \multicolumn{1}{c}{9.46}   & \multicolumn{1}{c}{12.19}    \\ 
& \multicolumn{1}{c|}{RMSE}  & \multicolumn{1}{c}{21.48}    
& \multicolumn{1}{c}{27.71}   & \multicolumn{1}{c}{33.48}    \\ \hline
\end{tabular}
\end{table}

 The method HA has a constant performance for all prediction horizons since it is based on log range information. Both HA and VAR fail to represent a good performance. They do not consider any trends, seasonality, or other factors that may influence the time series, making it less accurate for long-term forecasting or when dealing with volatile data patterns. Nonetheless, HA provides a simple and fast solution for simple forecasting tasks. AGCRN improves MAE performance compared to the previous two baselines but performs poorly on RMSE. The next two methods are transformer-based approaches. Informer lacks considering spatial dependencies since it only has attention and no graph convolution operations in its architecture. Whereas STTransformer employs both attention and graph convolution, so it performs better than Informer. Although transformers perform well on most sequential data, they are not the best choice for TSF, as it was also shown in the findings from \cite{zeng2023transformers}. In TSF the primary objective is to capture the temporal relationships between data points within an ordered sequence. Although positional encoding and token embeddings are utilized in Transformers to retain some level of ordering information for sub-series, the inherent nature of the permutation-invariant self-attention mechanism still leads to a loss of temporal information.

 MTGNN performs well on smaller prediction horizons. Its poor performance for 12 days prediction can be attributed to the fact that it employs Convolutional Neural Networks to understand temporal dependencies, while RNNs have proved to be a better choice for this purpose. GTS is the second-best-performing method after our proposed TransGlow, which shows the effectiveness of our Augmented encode-decoder architecture. Finally, our proposed TransGlow outperforms the other methods over all prediction horizons.

 \subsection{Complexity}

Table \ref{params} shows the total number of parameters for each method. While larger models may have the potential to achieve better performance on complex tasks, it is essential to balance model complexity with efficiency considerations to find the optimal trade-off between model performance and resource utilization. The total number of parameters can impact the efficiency of model training. During the inference phase, models with fewer parameters usually require less memory and computational power. Smaller models are often faster and more efficient in real-time applications. Of the two top-performing methods, GTS has the largest number of parameters, while our proposed TransGlow with the best MAE has third place in the parameters table. 

 \begin{table}[!htbp]
\begin{center}
\caption{Efficiency comparison of baseline methods}
\label{params}
\begin{tabular}{ c c  |c }
\hline

Method& & Number of Parameters \\
\hline
STTransformer& & 292,281\\ \hline
MTGNN& & 375,548\\  \hline
TransGlow & & 405,593\\  \hline
AGCRN& & 747,600\\  \hline
Informer& & 12,251,850\\  \hline
GTS& & 16,578,387\\  \hline
\end{tabular}
\end{center}
\end{table}

\section {Conclusion}
TransGlow is designed to continuously predict water flow over various monitoring stations along rivers, streams, and other water bodies. It can be integrated into flood warning systems to trigger alerts when the predicted water flow exceeds certain critical levels. Our proposed model can also help create flood hazard maps that identify areas at high risk of flooding based on predicted water flow patterns. 

 Water flow in a hydrological system is highly dependent on the interactions between different monitoring stations, such as rivers, streams, and lakes. GNNs can naturally capture these spatial dependencies by considering the graph structure of the network, which allows the model to incorporate the influence of nearby regions. GNNs can also effectively model temporal dynamics using recurrent units, such as the Graph Convolution Recurrent Neural Network (GCRN). Understanding the graph structure in the context of water systems can be challenging. It may not be static and can change over time due to various factors, such as weather conditions or human interventions. Graph learning methods can adaptively learn the connectivity between stations based on the available data. 

The encoder-decoder design is commonly used in prediction tasks. The benefit of the augmented attention layer proposed for the encoder-decoder architecture lies in its ability to focus selectively on relevant parts of the input sequence. The attention mechanism allows the decoder to access different parts of the encoded input sequence based on their importance for generating the output. Experiments on a real-world dataset show the advantage of our proposed model in terms of complexity and performance improvement.

Future works can explore integrating additional external factors influencing water flow, such as rainfall data, temperature, or land-use patterns. These factors can provide valuable context and improve the accuracy of the predictions. Investigating multi-task learning techniques to jointly predict other hydrological variables, such as water quality parameters or groundwater levels, alongside water flow can also lead to a more comprehensive understanding of the water system's behavior.

\section*{Acknowledgements}

C.P. - This research is supported by the Natural Sciences and Engineering Research Council of Canada Grants RGPIN-2021-03479 (Discovery Grant) and No. ALLRP 571887-2021 (Alliance Grant).\\
Z.P. - This research was undertaken, in part, based on support from the Gina Cody School of Engineering of Concordia University FRS.\\
U.E. - This research was undertaken, in part, thanks to funding from the Canada Excellence Research Chairs Program.

\bibliographystyle{IEEEtran}
\bibliography{IEEEabrv,IEEEexample}

\end{document}